\documentclass{article}

\PassOptionsToPackage{numbers, compress}{natbib}

 \usepackage[preprint]{nips_2018}

\usepackage[utf8]{inputenc} 
\usepackage[T1]{fontenc}    
\usepackage{hyperref}       
\usepackage{url}            
\usepackage{booktabs}       
\usepackage{amsfonts}       
\usepackage{nicefrac}       
\usepackage{microtype}      
\usepackage[cmex10]{amsmath}
\usepackage{amssymb}
\usepackage{graphicx}
\usepackage{soul,color}

\title{AttendNets: Tiny Deep Image Recognition Neural Networks for the Edge via Visual Attention Condensers}

\author{
  Alexander Wong$^{1,2,3,*}$, Mahmoud Famouri$^{3}$, and Mohammad Javad Shafiee$^{1,2,3}$\\
  $^{1}$ Vision and Image Processing Research Group, University of Waterloo, Waterloo, ON, Canada\\
  $^{2}$ Waterloo Artificial Intelligence Institute, University of Waterloo, Waterloo, ON, Canada\\
  $^{3}$ DarwinAI Corp., Waterloo, ON, Canada \\
  \texttt{$^{*}$ a28wong@uwaterloo.ca} \\
}

\begin{document}

\maketitle

\begin{abstract}
  While significant advances in deep learning has resulted in state-of-the-art performance across a large number of complex visual perception tasks, the widespread deployment of deep neural networks for TinyML applications involving on-device, low-power image recognition remains a big challenge given the complexity of deep neural networks.  In this study, we introduce \textbf{AttendNets}, low-precision, highly compact deep neural networks tailored for on-device image recognition.  More specifically, AttendNets possess deep self-attention architectures based on \textbf{visual attention condensers}, which extends on the recently introduced stand-alone attention condensers to improve spatial-channel selective attention.  Furthermore, AttendNets have unique machine-designed macroarchitecture and microarchitecture designs achieved via a machine-driven design exploration strategy.  Experimental results on ImageNet$_{50}$ benchmark dataset for the task of on-device image recognition showed that AttendNets have significantly lower architectural and computational complexity when compared to several deep neural networks in research literature designed for efficiency while achieving highest accuracies (with the smallest AttendNet achieving $\sim$7.2\% higher accuracy, while requiring $\sim$3$\times$ fewer multiply-add operations, $\sim$4.17$\times$ fewer parameters, and $\sim$16.7$\times$ lower weight memory requirements than MobileNet-V1).  Based on these promising results, AttendNets illustrate the effectiveness of visual attention condensers as building blocks for enabling various on-device visual perception tasks for TinyML applications. 
\end{abstract}

\section{Introduction}
\label{Introduction}
Deep learning~\cite{lecun2015deep} has resulted in significant breakthroughs in the area of computer vision, with state-of-the-art performance in a wide range of visual perception tasks such as image recognition~\cite{krizhevsky2012imagenet,ResNet,hu2017squeezeandexcitation}, object detection~\cite{fasterrcnn,liu2016ssd}, and semantic and instance segmentation~\cite{lin2017refinenet,chen2018deeplab,he2017mask,lin2019edgesegnet}.  Despite these breakthroughs, the widespread deployment of deep neural networks for tiny machine learning (TinyML) applications involving on-device visual perception on low-cost, low-power devices remains a major challenge given the increasing complexities of deep neural networks.  Motivated by the tremendous potential of deep learning empowering TinyML applications and inspired to tackle the aforementioned complexity challenge, there has been significant effort in recent years on the creation of highly efficient deep neural networks for edge scenarios.  These efforts in efficient deep learning have yielded a number of effective strategies, and can be typically grouped into two main categories: i) model compression, and ii) efficient architecture design.  In the realm of model compression, a popular approach is precision reduction~\cite{Jacob,Meng2017,courbariaux2015binaryconnect}, where the weights are represented at reduced data precision (e.g., fixed-point or integer precision~\cite{Jacob}, 2-bit precision~\cite{Meng2017}, 1-bit precision~\cite{courbariaux2015binaryconnect}).  Another model compression strategy leverage traditional data compression methods such as hashing, coding, and thresholding~\cite{han2015deep}.  More recently, teacher-student strategies for model compression has also been explored~\cite{hinton2015distilling,projectionnet}, where a larger teacher network is leveraged to train a smaller student network.

In the realm of efficient architecture design, a number of effective architecture design patterns geared around network efficiency have been introduced~\cite{MobileNetv1,MobileNetv2,SqueezeNet,SquishedNets,TinySSD,ShuffleNetv1,ShuffleNetv2,ResNet}.  One popular design pattern involves the introduction of bottlenecks~\cite{ResNet,MobileNetv2,SqueezeNet}, which serve the purpose of reducing dimensionality for more complex operation such as spatial convolutions.   Another efficient network design pattern involves the introduction of factorized convolutions~\cite{MobileNetv1,MobileNetv2}, which reduce architectural and computational complexity by factorizing spatial convolutions into smaller, more efficient operations.  A third efficient network design pattern involves the  introduction of new operations such as pointwise group convolutions and channel shuffling~\cite{ShuffleNetv1,ShuffleNetv2}.  Despite the range of strategies explored, one particular area that has not been well explored and is ripe for innovation is to leverage the concept of self-attention~\cite{bahdanau2014neural,vaswani2017attention,hu2017squeezeandexcitation,woo2018cbam,devlin2018bert}, seen as one of the recent big breakthroughs in deep learning, for the purpose of building highly efficient deep neural network architectures.

In this study, we introduce \textbf{AttendNets}, low-precision, highly compact deep neural networks tailored for on-device image recognition.  More specifically, AttendNets possess deep self-attention architectures based on \textbf{visual attention condensers}, which extends on the recently introduced stand-alone attention condensers to improve spatial-channel selective attention.  Furthermore, AttendNets have unique machine-designed macroarchitecture and microarchitecture designs achieved via a machine-driven design exploration strategy.  

The paper is organized as follows.  In Section~\ref{Methods}, we will describe in detail the concept of visual attention condensers and the machine-driven design exploration strategy leveraged to build the proposed AttendNets.  In Section~\ref{Architectures}, we will describe the produced AttendNet deep self-attention network architectures and discuss some interesting observations about their architecture designs.   In Section~\ref{Results}, we will present experimental results that quantitatively explore the efficacy of the proposed AttendNets when compared to previously proposed efficient deep neural networks for the task of on-device image recognition.  Fourth and finally, we will draw conclusions and discuss potential future directions in Section~\ref{Conclusions}.

\section{Methods}
\label{Methods}
\begin{figure}
\centering
	\includegraphics[width = \linewidth]{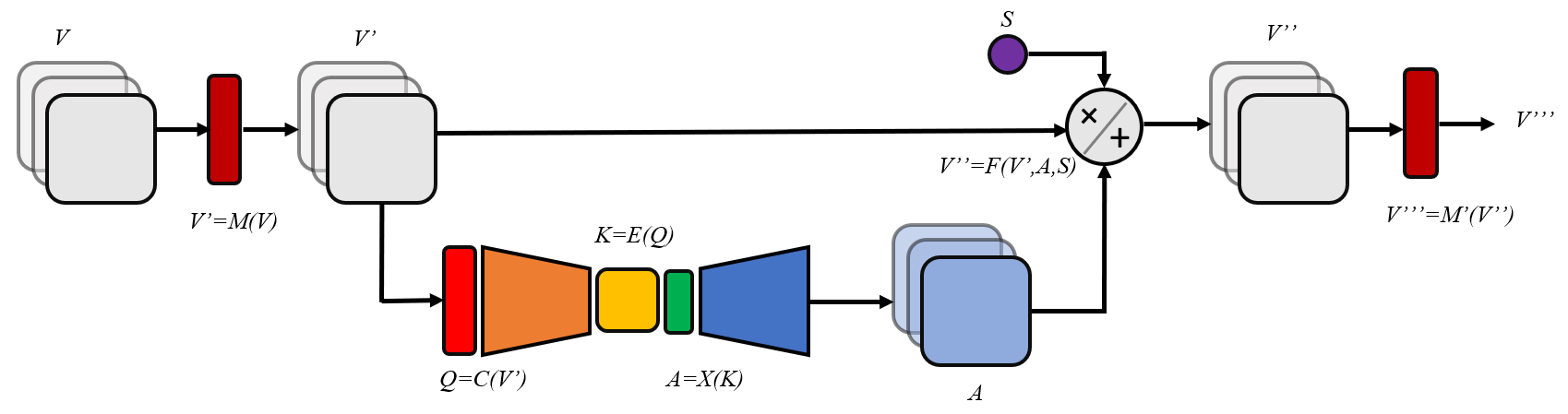}
	\caption{A visual attention condenser (VAC) is a self-attention mechanism consisting of a down-mixing layer, $M(V)$, condensation layer $C(V')$, an embedding structure $E(Q)$, an expansion layer $X(K)$, a selective attention mechanism $F(V',A,S)$, and an up-mixing layer $M'(V'')$.  The down-mixing layer $M(V)$ projects the input activations $V$ to a reduced channel dimensionality to $V'$.  The condensation layer $C(V')$ condenses $V'$ for reduced dimensionality to $Q$ with emphasis on strong activation proximity.  An embedding structure $E(Q)$ produces a condensed embedding $K$ from $Q$ characterizing joint spatial-channel activation relationships.  An expansion layer $X(K)$ projects the condensed embedding $K$ to higher dimensionality to produce self-attention values $A$.  The output $V''$ is a product of $V'$, self-attention values $A$, and scale $S$ via selective attention $F(V',A,S)$. Finally, the up-mixing layer $M'(V'')$ projects $V''$ to a higher channel dimensionality for final output $V'''$.}	
	\label{fig:condenser}
\end{figure}

In this study, we introduce \textbf{AttendNets}, a family of low-precision, highly compact deep neural networks tailored specifically for on-device image recognition.  Two key concepts are leveraged to construct the proposed AttendNets.  First, we introduce a stand-alone self-attention mechanism called \textbf{visual attention condensers}, which extends upon the recently introduced attention condensers~\cite{alex2020tinyspeech} to further improve its efficacy and efficiency for spatial-channel selective attention.  Second, we leverage a machine-driven design exploration strategy incorporating visual attention condensers to automate the network architecture design process to produce the proposed AttendNets in a way that strikes a strong balance between network complexity and image recognition accuracy.  Details pertaining to these two key concepts, along with details of the AttendNet network architectures, are provided below.

\subsection{Visual Attention Condensers}
The first concept we leverage to construct the proposed AttendNets is the concept of visual attention condensers.  The concept of self-attention in deep learning has led to significant advances in recent years~\cite{bahdanau2014neural,vaswani2017attention,hu2017squeezeandexcitation,woo2018cbam,devlin2018bert}, particularly with the advent of Transformers~\cite{vaswani2017attention} that has reshaped the landscape of machine learning for natural language processing.  It can be said that much of research on self-attention in deep learning has focused on improving accuracy, and this has had a heavy influence over the design of self-attention mechanisms.  For example, in the realm of computer vision, self-attention mechanisms have largely been explored as a mechanism for augmenting existing deep convolutional neural network architectures~\cite{hu2017squeezeandexcitation,woo2018cbam,bello2019attention}.  Many of the introduced self-attention mechanisms for augmenting deep convolutional neural network architectures have focused on the decoupling of attention into channel-wise attention~\cite{hu2017squeezeandexcitation} and local attention~\cite{woo2018cbam}, and are integrated to improve the selective attention capabilities of existing convolution modules to boost overall network accuracy at the expense of architectural and computational complexity.

Motivated to explore the design of self-attention mechanisms primarily in the direction of enabling more efficient deep neural network architectures instead of primarily for accuracy, Wong et al.~\cite{alex2020tinyspeech} introduce the concept of attention condensers as a stand-alone building block for deep neural networks geared around condensed self-attention.  More specifically, attention condensers are stand-alone self-attention mechanisms that learn and produce unified embeddings at reduced dimensionality for characterizing the joint cross-dimensional activation relationships.  The result is that attention condensers can better model activation relationships than decoupled mechanisms, resulting in improved selective attention while maintaining low complexity.  Furthermore, these improved modeling capabilities enable attention condensers to be leveraged as stand-alone modules, rather than as dependent modules geared for augmenting convolution modules.  As such, the heavier use of attention condensers and sparser use of complex stand-alone convolution modules can lead to more efficient deep neural network architectures while maintaining high modeling accuracy.  The efficacy of attention condensers as a means for building highly efficient deep neural networks was illustrated in~\cite{alex2020tinyspeech} for the task of on-device speech recognition, where deep attention condenser networks achieved computational and architectural complexities exceeding an order of magnitude when compared to previous deep speech recognition networks in literature.

Motivated by the promise of attention condensers, we extend upon the attention condenser design to further improve their efficiency and effectiveness for tackling visual perception tasks such as image recognition.  More specifically, we take inspiration from the observation that deep convolutional neural network architectures for tackling complex visual perception tasks often have very high channel dimensionality.  As such, while the existing attention condenser design can still achieve significant reductions on network complexity under such scenarios, we hypothesize that further complexity reductions can be gained through better handling of the high channel dimensionality when learning the condensed embedding of joint spatial-channel activation relationships.  As such, we introduce an extended visual attention condenser design where we introduce a pair of learned channel mixing layers that further reduces spatial-channel embedding dimensionality while preserving selective attention performance.

An overview of the proposed visual attention condenser (VAC) is shown in Figure~\ref{fig:condenser}.  More specifically, a visual attention condenser is a self-attention mechanism consisting of a down-mixing layer, $M(V)$, condensation layer $C(V')$, an embedding structure $E(Q)$, an expansion layer $X(K)$, a selective attention mechanism $F(V',A,S)$, and an up-mixing layer $M'(V'')$.  The down-mixing layer \mbox{$V'=M(V)$} learns and produces a projection of the input activations $V$ to a reduced channel dimensionality to obtain $V'$.  The condensation layer (i.e., $Q=C(V')$) condenses $V'$ for reduced dimensionality to $Q$ with an emphasis on strong activation proximity to better promote relevant region of interest despite the condensed nature of the spatial-channel representation.  An embedding structure (i.e., $K=E(Q)$) then learns and produces a condensed embedding $K$ from $Q$ characterizing joint spatial-channel activation relationships.  An expansion layer (i.e., $A=X(K)$) then projects the condensed embedding $K$ to an increased dimensionality to produce self-attention values $A$ emphasizing regions of interest in the same domain as $V'$.  The output $V''$ is a product of $V'$, self-attention values $A$, and scale $S$ via selective attention (i.e., $V''=F(V',A,S)$). Finally, the up-mixing layer $M'(V'')$ learns and prodcues a projection of $V''$ to a higher channel dimensionality for final output $V'''$ that has the same channel dimensionality as the input activation $V$.  Overall, through the introduction of the pair of learned mixing layers into the attention condenser design, a better balance between joint spatial-channel embedding dimensionality and selective attention performance can be achieved for building highly efficient deep neural networks for tackling complex visual perception problems on the edge.
\subsection{Machine-driven Design Exploration}

The second concept we leverage to construct the proposed AttendNets is the concept of machine-driven design exploration.  Similar to self-attention, the concept of machine-driven design exploration has gain tremendous research interests in recent years.  In the realm of machine-driven design exploration for efficient deep neural network architectures, several notable strategies have been proposed.  One strategy, which was taken by MONAS~\cite{MONAS}, ParetoNASH~\cite{ParetoNASH}, and MNAS~\cite{MNAS}, involves formulating the search of efficient deep neural network architectures as a multi-objective optimization problem, where the objectives may include model size, accuracy, FLOPs, device inference latency, etc. and a solution found via reinforcement learning and evolutionary algorithm.  Another strategy, which was taken by generative synthesis~\cite{Wong2018}, involves formulating the search of a generator of efficient deep neural networks as a constrained optimization problem, where the constraints may include model size, accuracy, FLOPs, device inference latency, etc. and a solution is found through an iterative solving process.  In this study, we leverage the latter strategy and leverage generative synthesis~\cite{Wong2018} to automate the process of generating the macroarchitecture and microarchitecture designs of the final AttendNet network architectures such that they are tailored specifically for the purpose of on-device image recognition in computational and memory constrained scenarios such as on low-cost, low-power edge devices, with an optimal balance between image recognition accuracy and network efficiency.

Briefly (the theoretical underpinnings of generative synthesis are presented in detail in~\cite{Wong2018}), generative synthesis formulates the following constrained optimization problem, which involves the search of a generator $\mathcal{G}$ whose generated deep neural network architectures $\left\{N_s|s \in S\right\}$ maximize a universal performance function $\mathcal{U}$ (e.g.,~\cite{Wong2018_Netscore}), with constraints around operational requirements defined by an indicator function $1_r(\cdot)$,
\begin{equation}
\mathcal{G}  = \max_{\mathcal{G}}~\mathcal{U}(\mathcal{G}(s))~~\textrm{subject~to}~~1_r(\mathcal{G}(s))=1,~~\forall s \in S.
\label{optimization}
\end{equation}
\noindent where $S$ denoting a set of seeds.  The approximate solution to this constrained optimization problem is found in generative synthesis through an iterative solving process, with initiation of this process based on a prototype $\varphi$, $\mathcal{U}$, and $1_r(\cdot)$.  As the goal for the proposed AttendNets is to achieve a strong balance between accuracy and network efficiency for the task of on-device image recognition, we explore an indicator function $1_r(\cdot)$ with two key constraints: i) the top-1 validation accuracy is greater than or equal to 71\% on the ImageNet$_{50}$ edge vision benchmark dataset introduced by Fang et al.~\cite{ImageNet50} for evaluating performance of deep neural networks for on-device vision applications, and ii) 8-bit weight precision.  First, a top-1 validation accuracy constraint of 71\% validation accuracy was chosen to make AttendNets comparable in accuracy to a state-of-the-art efficient deep neural network proposed in~\cite{alex2019attonets} for on-device image recognition.  Second, an 8-bit weight precision constraint was chosen to account for the memory constraints of low-cost edge devices.  Taking advantage of the fact that the generative synthesis process is iterative and produces a number of successive generators, we leverage two of the constructed generators at different stages to automatically generate two compact deep image recognition networks (AttendNet-A and AttendNet-B) with different tradeoffs between image recognition accuracy and network efficiency.

In terms of $\varphi$, we define a residual design prototype whose input layer takes in an RGB image, and the last layers consisting of global average pooling layer, a fully-connected layer, and a softmax layer indicating the image category.  How and where visual attention condensers should be leveraged is not defined in $\varphi$, and thus the macroarchitecture and microarchitecture designs of the final AttendNets is automatically by the machine-driven design exploration process to determine the best way to satisfy the specified constraints in $1_r(\cdot)$.

Finally, to realize the concept of visual attention condensers in a way that enables the learning of condensed embeddings characterizing joint spatial-channel activation relationships in an efficient yet effective manner, we leveraged max pooling, a lightweight two-layer neural network (grouped then pointwise convolution), unpooling, and pointwise convolution for the condensation layer $C(V')$, the embedding structure $E(Q)$, the expansion layer $X(K)$, and the mixing layers $M(V)$ and $M'(V'')$, respectively, within a visual attention condenser.

\begin{figure}
\centering
	\includegraphics[width = 1\linewidth]{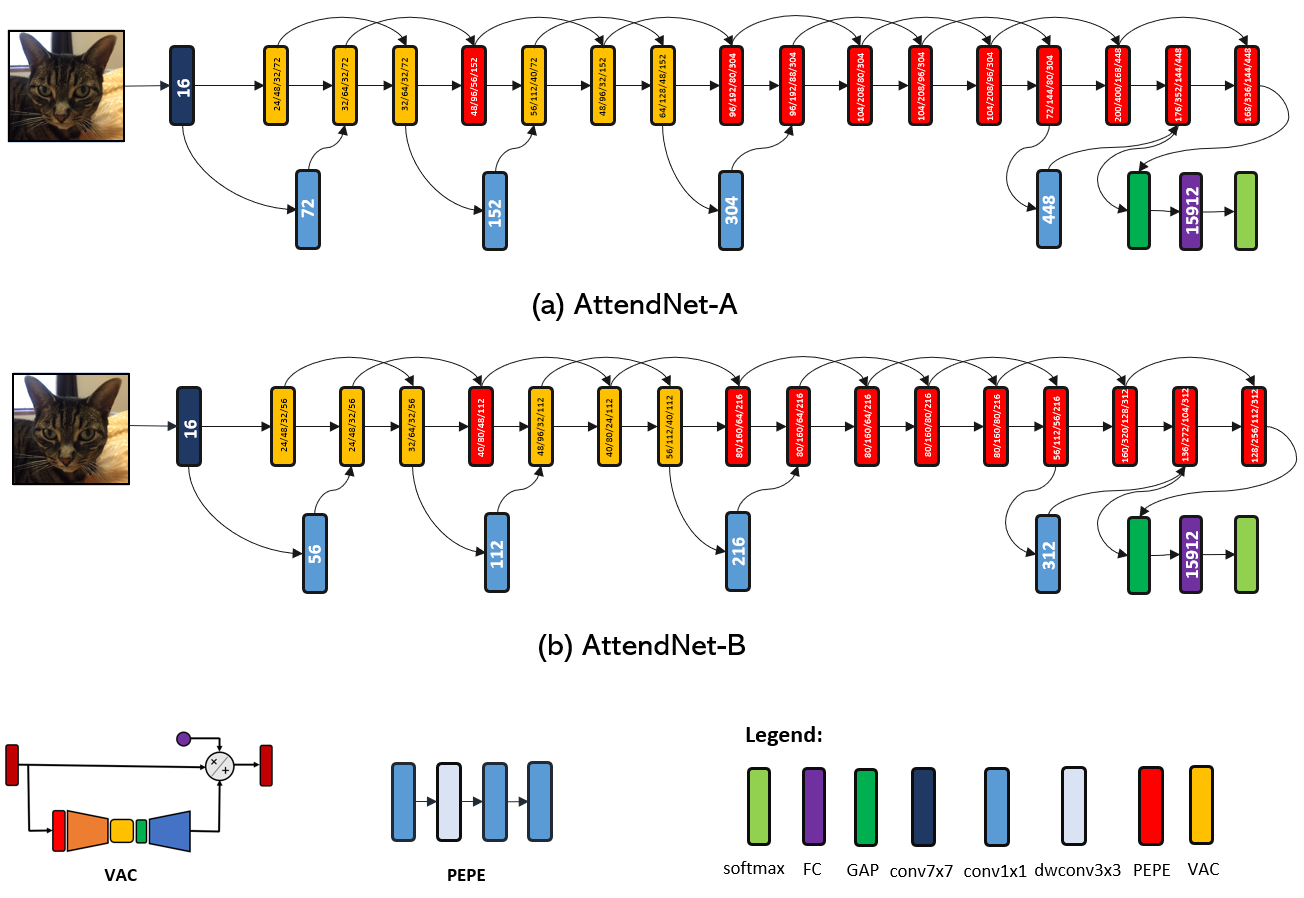}
	\caption{AttendNet architectures for image recognition.  The number in each convolution module represents the number of channels. The numbers in each visual attention condenser represents the number of channels for the down-mixing layer, the first and second layers of the embedding structure, and the up-mixing layer, respectively.  The numbers in each projection-expansion-projection-expansion (PEPE) module represents the number of channels for the first expansion layer, first expansion layer, second projection layer, and second projection layer, respectively.  The number in each fully-connected layer represents the number of synapses.  An image acts as the input to AttendNet network architectures, which comprises of a convolutional layer, followed by a mix of consecutive visual attention condensers (VACs) that perform consecutive visual selective attention and PEPE modules for efficient feature representation.  The final output feed into a global average pooling layer, a fully-connected layer, and finally a softmax layer.  The AttendNet architectures exhibit high architectural diversity both at the macroarchitecture and microarchitecture level as a result of the machine-driven design exploration, and heavy use of visual attention condensers early on in the network architecture while relying more on PEPE modules later in the network architecture.}	
	\label{fig:attendnets}
\vspace{-0.21 cm}
\end{figure}

\section{AttendNet Architecture Designs}
\label{Architectures}
Figure~\ref{fig:attendnets} shows the architecture designs of the two AttendNets, produced via the aforementioned machine-driven design exploration strategy that incorporates visual attention condensers in its design considerations.  A number of interesting observations can be made about the AttendNet architecture designs.  First, it can be observed that the AttendNet architecture designs are comprised of a mix of consecutive stand-alone visual attention condensers performing consecutive visual selective attention and projection-expansion-projection-expansion (PEPE) modules for efficient feature representation.  The PEPE module was discovered by the machine-driven exploration strategy and comprises of a projection layer that reduces dimensionality via pointwise convolution, an expansion layer that increases dimensionality efficiently via depthwise convolution, a projection layer that reduces dimensionality again via pointwise convolution, and finally an expansion layer that increases dimensionality again via pointwise convolution.  

Second, it can be observed that there is a heavy use of visual attention condensers early on within the AttendNet architecture by the machine-driven design exploration strategy, while relying on PEPE modules later in the network architecture.  This interesting design choice by the machine-driven design exploration strategy may be a result of selective attention being more important earlier on low-level to medium-level visual abstraction for image recognition to enable better focus on irrelevant regions of interest critical to decision-making within a complex scene. 

Third and finally, it can be observed that the AttendNet network architectures exhibits high architectural diversity both at the macroarchitecture level and microarchitecture level.  For example, at the macroarchitecture level, there is a heterogeneous mix of visual attention condensers, PEPE modules, spatial and pointwise convolutions, and fully-connected layers.  At the microarchitecture level, the visual attention condensers and PEPE modules have a diversity of microarchitecture designs as seen by the differences in channel configurations.  This level of architectural diversity is a result of the machine-driven design exploration process, which has the benefit of determining the best architecture design at a fine grained level to achieve a strong balance of network efficiency and accuracy for the specific task at hand.  Based on these three interesting observations, it can be seen the proposed AttendNet network architectures is highly tailored for on-device image recognition for the edge, and also shows the merits of leveraging both visual attention condensers and machine-driven design exploration for achieving such highly efficient, high-performance deep neural networks. 

\section{Results and Discussion}
\label{Results}
\begin{table}[h]
	\centering
	\caption{Top-1 accuracy, number of parameters, and number of multiply-add operations of AttendNets in comparison to four efficient deep image recognition networks (MobileNet-V1~\cite{MobileNetv1}, MobileNet-V2~\cite{MobileNetv2}, AttoNet-A~\cite{alex2019attonets}, AttoNet-B~\cite{alex2019attonets}).  Best results are in \textbf{bold}. Results for AttendNets  based on 8-bit low precision weights, while results for other tested networks based on 32-bit full precision weights}
	\begin{tabular}{p{3cm}cc|cc}
		\hline
		\textbf{Model}& \textbf{Top-1 Accuracy} & \textbf{Params} &	\textbf{Mult-Adds} \\
		\hline 
		MobileNet-V1~\cite{MobileNetv1}	&	$64.5\%$	&	3260K	&	567.5M \\
		MobileNet-V2~\cite{MobileNetv2}		&	$68.7\%$	&	2290K	&	299.7M \\
		AttoNet-A~\cite{alex2019attonets} 	&      73.0\% 	&	2970K 	&	424.8M \\
		AttoNet-B~\cite{alex2019attonets} 	&	71.1\% 	&	1870K 	&	277.5M \\
		\hline
		AttendNet-A		& 	\textbf{73.2\%}	&	1386K	&	276.8M \\
		AttendNet-B		& 	71.7\%	&	\textbf{782K}	&	\textbf{191.3M} \\
		\hline
	\end{tabular}\\
	\label{tab_Results}
\end{table}	

In this study, we evaluate the efficacy of the proposed low-precision AttendNets on the task of image recognition to empirically study the balance between accuracy and network efficiency.  More specifically, we leverage ImageNet$_{50}$, a benchmark dataset that was introduced by Fang et al.~\cite{ImageNet50} for evaluating performance of deep neural networks for on-device vision applications on the edge  derived from the popular ImageNet~\cite{ImageNet} dataset.  To quantify accuracy and network efficiency, we computed the following performance metrics: i) top-1 accuracy, ii) the number of parameters (to quantify architectural complexity), and iii) the number of multiply-add operations (to quantify computational complexity).  For comparative purposes, the same performance metrics were also evaluated on MobileNet-V1~\cite{MobileNetv1}, MobileNet-V2~\cite{MobileNetv2}, AttoNet-A~\cite{alex2019attonets}, and AttoNet-B~\cite{alex2019attonets}, four highly efficient deep image recognition networks that were all designed for on-device image recognition purposes.

Table~\ref{tab_Results} shows the top-1 accuracy, the number of parameters, and the number of multiply-add operations of the AttendNets alongside the four other tested efficient deep image recognition networks.  From the quantitative results, it can be clearly observed that the proposed AttendNets achieved a significantly better balance between accuracies and architectural and computational complexity when compared to the other tested efficient deep neural networks.  In terms of lowest architectural and computational complexity, AttendNet-B achieved significantly higher accuracy compared to MobileNet-V1 ($\sim$\textbf{7.2\%} higher) but requires $\sim$\textbf{4.17$\times$} fewer parameters, $\sim$\textbf{16.7$\times$} lower weight memory requirements, and $\sim$\textbf{3$\times$} fewer multiply-add operations than MobileNet-V1.  Compared to similarly-accurate state-of-the-art AttoNet-B, AttendNet-B achieved $\sim$\textbf{0.6\%} higher accuracy but requires $\sim$\textbf{2.4$\times$} fewer parameters, $\sim$\textbf{9.6$\times$} lower weight memory requirements, and $\sim$\textbf{1.45$\times$} fewer multiply-add operations.  In terms of the highest top-1 accuracy, AttendNet-A achieved significantly higher accuracy compared to MobileNet-V1 and MobileNet-V2 ($\sim$\textbf{8.7\%} higher and $\sim$\textbf{4.5\%} higher, respectively)  but requires $\sim$\textbf{2.35$\times$} fewer parameters, $\sim$\textbf{9.4$\times$} lower weight memory requirements, and $\sim$\textbf{2.1$\times$} fewer multiply-add operations than MobileNet-V1 and $\sim$\textbf{1.65$\times$} fewer parameters, $\sim$\textbf{6.6$\times$} lower weight memory requirements, and $\sim$\textbf{1.1$\times$} fewer multiply-add operations than MobileNet-V2.  Compared to the similarly accurate state-of-the-art AttoNet-A, AttendNet-A achieved slightly higher accuracy ($\sim$\textbf{0.2\%} higher) but requires $\sim$\textbf{2.1$\times$} fewer parameters, $\sim$\textbf{8.4$\times$} lower weight memory requirements, and $\sim$\textbf{1.53$\times$} fewer multiply-add operations.

These quantitative performance results illustrate the efficacy of leverage both visual attention condensers and machine-driven design exploration to creating highly-efficient deep neural network architectures tailored for on-device image recognition that striking a strong balance between accuracy and network complexity.
\section{Conclusions}
\label{Conclusions}
In this study, we introduce \textbf{AttendNets}, low-precision, highly compact yet high-performance deep neural networks tailored for on-device image recognition.  The strong balance between accuracy and network efficiency achieved by AttendNets was achieved through the introduction of visual attention condensers, stand-alone self-attention mechanisms that extend upon the concept of attention condensers to improve spatial-channel selective attention in an efficient manner, and the use of a machine-driven design exploration strategy to automatically determine the macroarchitecture and microarchitecture designs of the AttendNet deep self-attention architectures.  We demonstrated the efficacy of the resulting AttendNets on the task of on-device image recognition, which were able to achieve significantly better balance between accuracy and network efficiency when compared to previously proposed efficient deep neural networks designed for on-device image recognition.  As such, the visual attention condensers used to create AttendNets have the potential to be a useful building block for constructing highly efficient, high-performance deep neural networks for real-time visual perception tasks on low-power, low-cost edge devices for TinyML applications.

Given the promising results associated with AttendNets and the use of visual attention condensers to achieve highly efficient yet high-performance deep neural networks, future work involves exploring the effectiveness of AttendNets on downstream tasks such as object detection, semantic segmentation, and instance segmentation to empower a wider variety of vision-related TinyML applications ranging from autonomous vehicles to smart city monitoring to wearable assistive technologies to remote sensing.  We also aim to explore different design choices for the individual components of the visual attention condenser (e.g., mixing layers, embedding structure, condensation layer, expansion layer) and their impact on accuracy and efficiency.  Finally, as alluded to in~\cite{alex2020tinyspeech}, the exploration of self-attention architectures based on visual attention condensers and their adversarial robustness is a worthwhile endeavor, particularly given recent focuses on robustness and dependability of deep learning.

\section{Broader Impact}
\label{BroaderImpact}
There has been tremendous recent interest in TinyML (tiny machine learning) as being one of the key disruptive technologies towards widespread adoption of machine learning in industry and society. By enabling improved automated decision-making and predictions via machine learning to operate directly and in real-time on low-cost, low-power embedded hardware such as microcontrollers and embedded microprocessors, TinyML facilitates a wide variety of applications ranging from smart homes to smart factories to smart cities to smartgrids where a large number of intelligent nodes work in unison to improve productivity, efficiency, consistency, and performance. Furthermore, in mission-critical scenarios where privacy and security is crucial, such as in automotive, security, and aerospace, TinyML facilitates tetherless intelligence without the requirement of continuous connectivity, thus enabling greater dependability.  As such, TinyML can have a tremendous impact across all facets of society and industry and thus have important socioeconomic implications that needs to be considered.  In our exploration of new strategies such as the proposed visual attention condensers in this study, the hope is that the insights gained from such explorations can be leveraged by the community to advance efforts in TinyML for greater adoption of machine learning as a ubuiquitous technology.

\footnotesize
\bibliographystyle{IEEEtran}
\bibliography{tinyvision}

\end{document}